\documentclass{article}

\PassOptionsToPackage{round,authoryear}{natbib}

\usepackage[preprint]{neurips_2026}

\usepackage{amsmath,amsfonts,bm}

\def\eqref#1{equation~\ref{#1}}
\def\1{\bm{1}}

\def\mX{{\bm{X}}}

\DeclareMathAlphabet{\mathsfit}{\encodingdefault}{\sfdefault}{m}{sl}
\SetMathAlphabet{\mathsfit}{bold}{\encodingdefault}{\sfdefault}{bx}{n}
\newcommand{\tens}[1]{\bm{\mathsfit{#1}}}

\def\tX{{\tens{X}}}

\newcommand{\R}{\mathbb{R}}

\usepackage{amsmath}
\usepackage{graphicx}
\usepackage{multirow}
\usepackage{enumitem}
\usepackage{wrapfig}
\usepackage[flushleft]{threeparttable}
\usepackage{makecell}
\usepackage[utf8]{inputenc}
\usepackage[T1]{fontenc}
\usepackage{hyperref}
\usepackage{url}
\usepackage{booktabs}
\usepackage{amsfonts}
\usepackage{nicefrac}
\usepackage{microtype}
\usepackage{xcolor}
\usepackage{subfig}
\usepackage{xspace}
\usepackage{array}
\usepackage{diagbox}
\usepackage{algorithmicx,algorithm}
\usepackage{subcaption}

\usepackage{amssymb}
\usepackage{mathtools}
\usepackage{amsthm}
\usepackage[normalem]{ulem}
\usepackage{tabularx}
\usepackage{color}
\usepackage{xspace} 
\usepackage{diagbox}
\usepackage{balance}
\usepackage{marvosym}

\PassOptionsToPackage{numbers, compress}{natbib}

\title{Hermes: A Multi-Scale Spatial-Temporal Hypergraph Network for Stock Time Series Forecasting}

\author{Xiangfei Qiu$^{1}$, Liu Yang$^{1}$, Xiangyu Xu$^{1}$, Hanyin Cheng$^{1}$, Xingjian Wu$^{1}$, Rongjia Wu$^{2}$, \\\textbf{Zhigang Zhang$^{2}$, Ding Tu$^{2}$, Chenjuan Guo$^{1}$, Bin Yang$^{1}$, Christian S. Jensen$^{3}$, Jilin Hu$^{1}$ }\\$^1$\textit{East China Normal University, Shanghai, China}\\ $^2$\textit{CFETS Information Technology (Shanghai) Co., Ltd, Shanghai, China}  \\ $^3$\textit{Aalborg University, Aalborg, Denmark} 
}

\begin{document}

\maketitle

\begin{abstract}
Time series forecasting occurs in a range of financial applications providing essential decision-making support to investors, regulatory institutions, and analysts. Unlike multivariate time series from other domains, stock time series exhibit industry correlation. Exploiting this kind of correlation can improve forecasting accuracy. However, existing methods based on hypergraphs can only capture industry correlation relatively superficially. These methods face two key limitations: they do not fully consider inter-industry lead-lag interactions, and they do not model multi-scale information within and among industries. This study proposes the \textbf{Hermes} framework for stock time series forecasting that aims to improve the exploitation of industry correlation by addressing these limitations. The framework integrates moving aggregation and  multi-scale fusion modules in a hypergraph network. Specifically, to more flexibly capture the lead-lag relationships among industries, Hermes proposes a hyperedge-based moving aggregation module. This module incorporates a sliding window and utilizes dynamic temporal aggregation operations to consider lead-lag dependencies among industries. Additionally, to effectively model multi-scale information, Hermes employs cross-scale, edge-to-edge message passing to integrate information from different scales while maintaining the consistency of each scale. Experimental results on multiple real-world stock datasets show that Hermes outperforms existing state-of-the-art methods. 
\end{abstract}

\section{Introduction}
Stock time series are typically composed of data from multiple stocks, each of which includes multiple key indicators such as stock opening prices, closing prices, etc.~\cite{StockMixer,li2024master}. Stock time series forecasting~(STSF) is important in financial applications, providing crucial decision-making support for investors, regulatory institutions, and analysts, helping them make data-driven decisions in complex and ever-changing market environments~\cite{rather2017stock,faloutsos2018forecasting,jacob1999fintime,zheng2023relational}. 

\begin{figure*}[t]
  \centering
  \subfloat[Intra-industry correlation]
  {\includegraphics[width=0.323\textwidth, height=0.15\textheight]{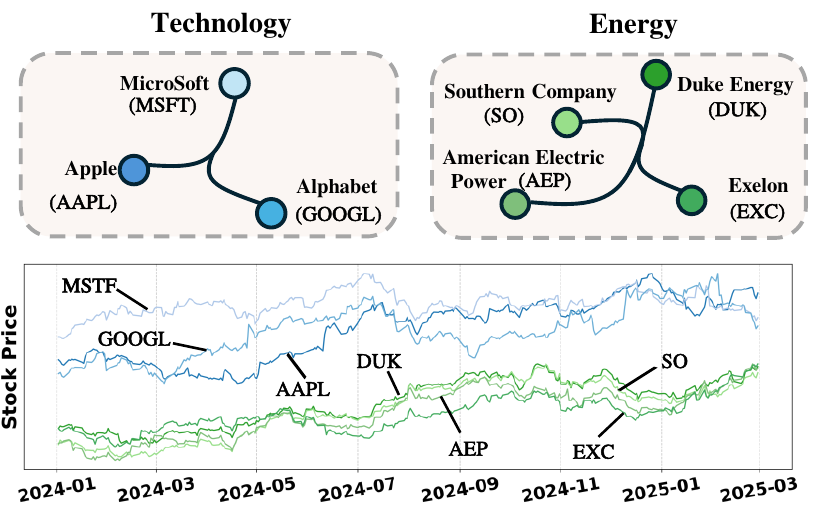}\label{Intra-industry correlation.}}
  \hspace{0.5mm}\subfloat[Lead-lag correlation]
  {\includegraphics[width=0.326\textwidth, height=0.145\textheight]{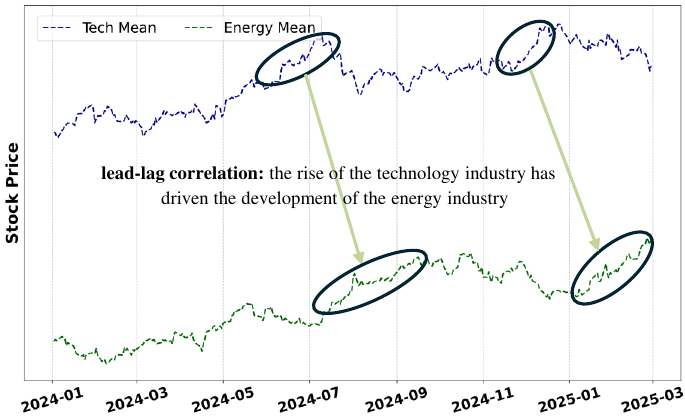}\label{Lead-lag correlation}}
  \hspace{0.5mm}\subfloat[ Multi-scale correlation]
  {\includegraphics[width=0.329\textwidth, height=0.145\textheight]{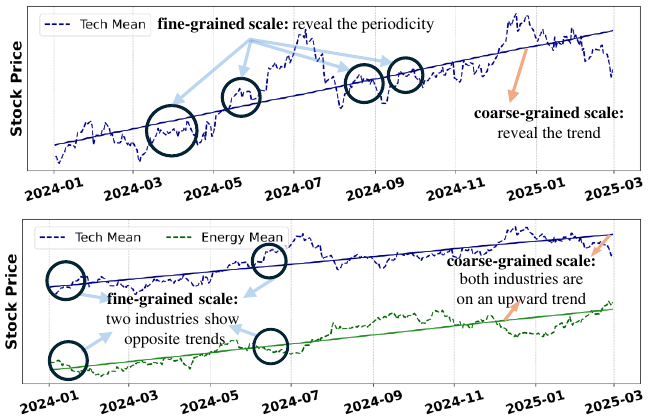}\label{Multi-scale correlation}}
  \caption{(a): Intra-industry correlation. Relations between stocks within an industry (hyperedge) can be
represented by a hypergraph. (b): Lead-lag correlation. (c):  Multi-scale correlation.}
    \label{Hyperparameter.}
    % \vspace{-10mm}
\end{figure*}

Most existing studies treat stocks in stock time series independently, thus disregarding the interconnected nature of market dynamics and contradicting the intrinsic interdependencies observed in real-world financial systems~\cite{diebold2014network}. In reality, stocks are often interrelated, and there are rich signals in the relationships between them~\cite{nobi2014effects}. For example, with the breakthroughs of Artificial Intelligence technology, companies like Apple, Microsoft, and Alphabet that actively invest and innovate in this domain are expected to lead in the market due to their technological advantages. It is one of the important factors for their stock prices to show a steady upward trend in the overall stock market\footnote{\tiny
\nolinkurl{www.morningstar.co.uk/uk/news/258865/tech-and-communication-stocks-drove-us-market-gains-in-2024.aspx}}
---see Figure~\ref{Intra-industry correlation.}. 
At the same time, the soaring power demand, driven by electrification, decarbonization, and the advent of artificial intelligence, led to a concurrent rise in the stock prices of companies in the energy sector\footnote{{\tiny\nolinkurl{https://finance.yahoo.com/news/american-electric-power-aep-among-121752028.html?}\label{2}}}---see Figure~\ref{Intra-industry correlation.}. Such industry correlations have gradually attracted more attention. Especially within the same industry, different stocks often exhibit correlations, reflecting the common economic environment, market fluctuations, policy changes, and other factors that impact the stocks. 

% Therefore, how to consider the mutual relationships among stocks within the same industry can enhance the modeling and forecasting capabilities of industry dynamics.

Hypergraphs can connect multiple graph nodes (representing, e.g., stocks) through hyperedges. By connecting stocks within the same industry by hyperedges to construct a hypergraph network, it is possible to capture more comprehensively the interrelationships and dynamics among stocks within an industry. However, existing hypergraph-based STSF methods that consider industry correlations still have the following challenges. 

\textbf{Challenge 1: Current methods fail to fully consider inter-industry lead-lag interactions.} In financial markets, there are time dependencies among industries, and these relationships are highly diverse and complex. As shown in Figure \ref{Lead-lag correlation}, the technology industry has flourished with breakthroughs in key technologies and market innovations, achieving significant progress. Technological innovations from technology companies have injected new vitality into the development of renewable energy, and energy companies have leveraged technologies such as big data, AI, and the Internet of Things to enhance the efficiency of energy production and management\footref{2}, thereby driving up stock prices~\cite{abid2025robust,olanrewaju2024strategic,rafiei2025comprehensive}. Therefore, in the modeling process, it is important to capture lead-lag correlations among industries more accurately, so as to better reflect the dynamic interactions and dependencies in financial markets, ultimately improving the accuracy and practicality of the forecasts.

\textbf{Challenge 2: Current methods do not model multi-scale information within and among industries.} Stock time series inherently possess multi-scale characteristics, and markets exhibit different features and patterns at different time scales. For example, in Figure~\ref{Multi-scale correlation}, we can observe a clear overall trend at a coarse-grained scale, and a cyclical pattern at a fine-grained scale. Moreover, two different industries, may exhibit correlations at a coarse-grained scale. However, delving deeper into a fine-grained analysis might find that two industries show opposite trends. Single-scale analysis methods cannot fully capture such characteristics, so there is a pressing need to adopt multi-scale analysis to reveal the dynamic changes in financial markets across different time scales, thereby providing richer and more accurate information for stock time series forecasting.

% In the short term, markets may be influenced by sudden events, investor sentiment, and other factors, resulting in significant volatility, while in the long term, macroeconomic factors and policy regulations play a more significant role.

In this study, we address the two challenges by proposing a general stock time series forecasting framework, \textbf{Hermes}, which integrates \textit{hyperedge-based moving aggregation} and \textit{hyperedge-based multi-scale fusion} modules into a spatial-temporal hypergraph network. These integrations enhance the precision and stability of stock forecasting. First, to more flexibly model lead-lag relationships among industries, we integrate a \textit{hyperedge-based moving aggregation module} into the hypergraph network. This module incorporates a sliding window and utilizes dynamic temporal aggregation operations to consider lead-lag dependencies among industries. This approach aims to capture a market's complex structure more precisely. Second, to address the issue that existing methods fail to effectively model multi-scale information, we decompose raw data at multiple scales and integrate a \textit{multi-scale fusion module} into the hypergraph network. This module employs cross-scale, edge-to-edge message passing to integrate information from different scales while preserving the consistency of each scale, thereby boosting both the accuracy and stability of forecasts. Experimental results on multiple real-world stock datasets show that Hermes outperforms existing state-of-the-art methods.

Our contributions are summarized as follows.
\begin{itemize}[left=0.5cm]
\item To solve the STSF problem, we propose a general spatial-temporal hypergraph network framework, \textbf{Hermes}, which learns more accurate and adaptive forecasting models by considering lead-lag and multi-scale industry correlations.

\item We propose a hyperedge-based moving aggregation module that introduces a sliding window and utilizes dynamic temporal aggregation operations to consider lead-lag dependencies among industries.

\item We design a hyperedge-based multi-scale fusion module. This module utilizes a cross-scale, edge-to-edge message-passing to integrate information from different scales while maintaining the consistency of each scale.

\item We report on extensive experiments on public stock datasets, finding that Hermes is capable of outperforming state-of-the-art baselines. 

\end{itemize}

\section{Related Work}
\label{Related Works}
\subsection{Deep Learning Stock Time Series Forecasting}
% While deep learning has made impressive advances in NLP and CV~\cite{zhang2025imdprompter,zhang2024can,zhang2025rethinking,chen2024gim,zhang2024distilling}, demonstrating that learned features can outperform human-designed features, they also evolve over time series. 
With the rise of deep learning, notable progress has occurred in stock forecasting. This is mainly because deep neural networks can model complex nonlinear patterns.  Some studies use recurrent neural networks (RNNs)~\cite{nelson2017stock,qin2017dual,akita2016deep,rahman2019predicting}  to capture the historical development in individual stock prices and predict their short-term trends. In addition, due to the significant progress of Multi-Layer Perceptrons (MLPs) architecture in the general time series field~\cite{zeng2023transformers,lin2025cyclenet,SparseTSF}. Increasingly many proposals target stock forecasting performance by improving the MLPs architecture. For example, a series of studies~\cite{StockMixer,lazcano2024back,tashakkori2024forecasting} enhance the learning ability of simple MLPs structures by utilizing the MLP-Mixer backbone, thereby improving forecasting accuracy. Next, the Transformer model has become increasingly popular due to the ability of its self-attention mechanism to capture long-term dependencies. For example, Master~\cite{li2024master} proposes a novel stock transformer for stock price forecasting to effectively capture stock correlation. While these deep learning methods have advanced stock forecasting, the complexity of stock data and the need to account for intricate relationships among stocks have led researchers to explore new methods based on hypergraphs.

\subsection{Hypergraph-based Stock Time Series Forecasting}
Traditional graphs have limitations when it comes to modeling higher-order multivariate relationships, as they can only represent relations between pairs of nodes. To address this issue, Hypergraph Neural Networks (HGNNs)~\cite{feng2019hypergraph} were introduced. Hypergraphs extend the concept of simple graphs to enable the capture of relationships among multiple stocks~\cite{chen2020hypergraph}. Hypergraphs have gained increasing attention and application across various fields. Hyperedge in a hypergraph represents a set of vertices, making them suitable for modeling non-pairwise relationships~\cite{luo2014stock,sawhney2020spatiotemporal,huynh2023efficient}. For example, HGAM~\cite{li2022hypergraph} is a hypergraph-based reinforcement learning method for stock portfolio selection. THGNN~\cite{xiang2022temporal} is a temporal and heterogeneous graph neural network model that aims to learn dynamic relationships by using two-stage attention mechanisms. STHAN-SR~\cite{sawhney2021stock} enhances corporate relevance based on Wiki data and uses hypergraph convolution to propagate information from higher-order neighbors. ESTIMATE~\cite{huynh2023efficient}, employs hypergraphs to capture non-pairwise correlations, utilizing temporal generative filters to identify individual patterns for each stock. Unlike the above methods, we propose a new hypergraph-based stock time series forecasting model that specifically takes into account the multi-scale nature of stock data, enabling it to capture the complexity of market dynamics more comprehensively. Furthermore, our approach places particular emphasis on the lead-lag relationships between financial industries, where the changes in one industry may precede or lag behind those in another. By capturing these key relationships more fully, we aim to improve forecasting accuracy.

\begin{figure*}[t]
    \centering
    \includegraphics[width=1\linewidth]{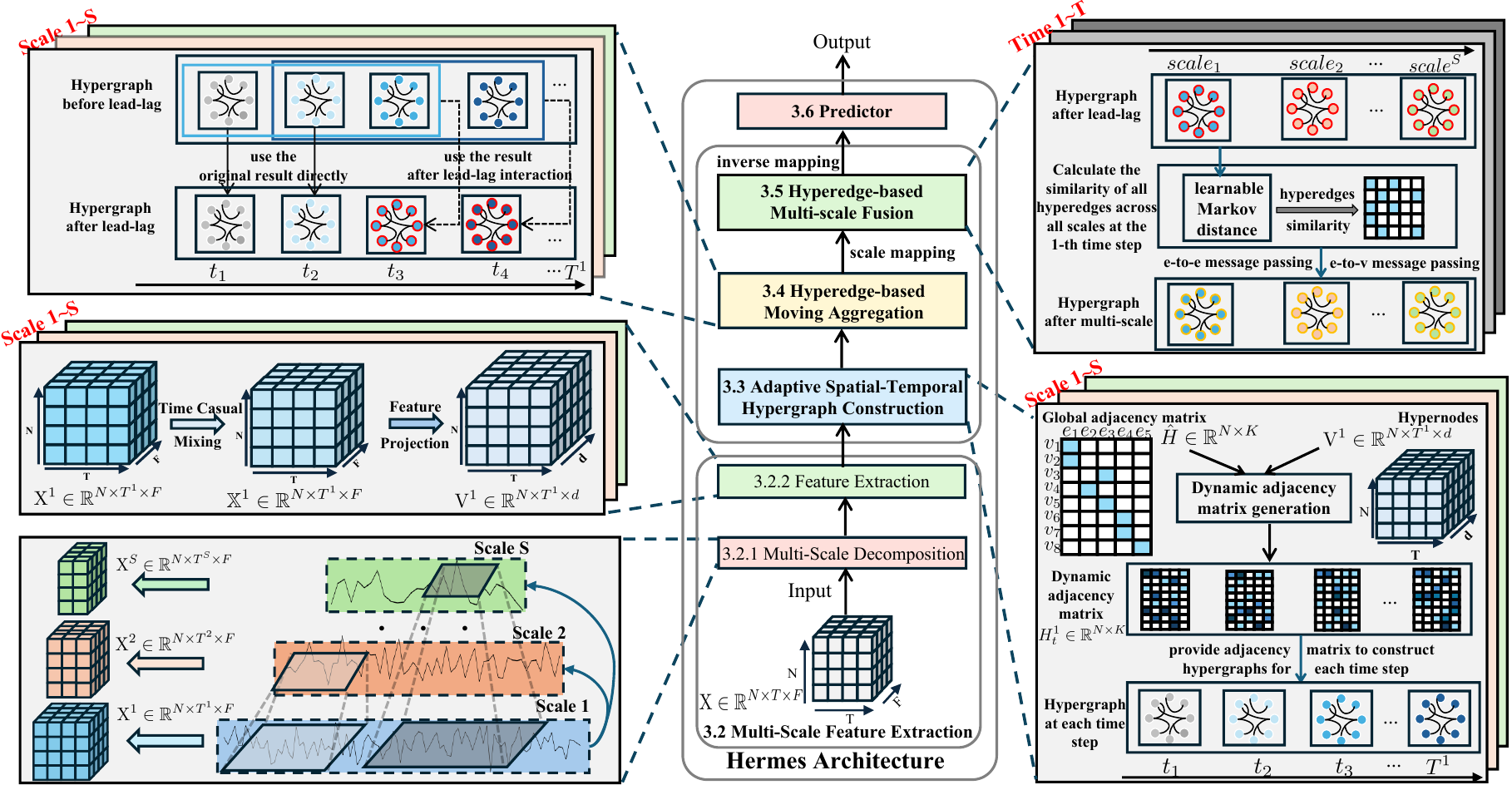}
    \caption{The architecture of Hermes.}
    \label{fig:arch}
% \label{industry correlation.}
\end{figure*}

% \begin{figure*}[t]
%     \centering
%     \includegraphics[width=1\linewidth]{figures/main1.pdf}
%     \caption{The architecture of Hermes.}
%     \label{fig:arch}
% % \label{industry correlation.}
% \end{figure*}

\section{Methodology}
\label{Methodology}
In financial markets, a stock time series $\tX =\left\{\mX_{1},\ldots,\mX_{N}\right\}$ records the historical data of $N$ stocks, where $\mX_i \in \R^{T \times F}$ is data of one stock with a lookback window of $T$ time steps and with $F$ technical indicators at each time step. In stock time series, the indicators may include opening price, closing price, highest price, lowest price, and trading volume. Following existing studies~\cite{huynh2023efficient,StockMixer}, we input a stock time series with multiple indicators, denoted as $\tX \in \mathbb{R}^{N \times T \times F}$, and aim to predict the closing price $p_i^t$ at time step $t$. We define the 1-day return ratio as $r_i^t = \frac{p_i^t - p_{i}^{t-1}}{p_{i}^{t-1}}$. With $\theta$ representing the model parameters, the process can be expressed as follows.
\begin{align}
\tX \in \mathbb{R}^{N \times T \times F} \stackrel{\theta}{\rightarrow} p \in \mathbb{R}^{N \times 1} \rightarrow r \in \mathbb{R}^{N \times 1}.
\end{align}

\subsection{Framework Overview}
In this section, we introduce the framework of our proposed spatiotemporal hypergraph neural network that considers multiple time scales and models the lead-lag relationships between industries---see Figure~\ref{fig:arch}. Our framework takes historical financial sequences as input, first samples the time series into several sequences of different granularities according to different time scales, and maps the feature dimension to the hidden dimension to consider the high-dimensional correlations between time series segments under different scales. Subsequently, we consider constructing relationships between stocks at each scale, specifically considering constructing hyperedges for stocks in the same industry, and modeling the lead-lag relationships between industries in the corresponding hypergraphs. Then, we model the interactions between different scales, construct hyperedges for industries of different scales, and adaptively learn the correlations between them under the corresponding multi-scale hypergraph, thereby playing a role in fusing scales. Finally, we predict the corresponding indicators for each stock for the next day through a prediction head based on a multilayer perceptron.
\subsection{Multi-Scale Feature Extraction}
\label{Multi-Scale Decomposition}
\noindent
\textbf{Multi-Scale Decomposition.}
Time series of different scales exhibit different characteristics, among which fine scales primarily describe detailed patterns and contain more periodic information, while coarse scales highlight macro changes and cover the overall trend changes. This multi-scale perspective can essentially unravel the complex variations within multiple components, thereby facilitating time variation modeling. It is worth noting that, especially for prediction tasks, due to their different dominant time patterns, multi-scale time series show different predictive capabilities.

Specifically, for the financial input time series \(\tX \in \mathbb{R}^{N \times T \times F}\), which denotes the records of $F$ key indicators for $N$ stocks of $T$ timestamps in the past. To decouple the intricate multi-resolution information, we use convolutional layers of different sizes to adaptively down-sample the time series into sequences of different scales:
\begin{align}
    \tX^i = \text{1D-Conv}^i(\tX),
\end{align}
\noindent where $\tX^i\in \mathbb{R}^{N \times T^i \times F}$ denotes the $i$-th scale of financial time series, $T^i$ is the down-sampled length. We then obtain the multi-scale time series sequences $\{\tX^1, \tX^2, \cdots, \tX^S\}$ of $S$ kinds of scales through $S$ convolutional layers with different kernel sizes and stride lengths.

\noindent
\textbf{Feature Extraction.}
Subsequently, for time series of different scales, we enhance their representations at two aspects. First, we adopt the Causal-Mixing technique to process the temporal dimension. The Causal-Mixing technique ensures causal relationships rather than full connectivity, and ensure that time series points can only see previous points. It consists of a group of MLP layers:
\begin{gather}
    \text{Causal-Mixing}^i = \{\text{MLP}^{i,1},\text{MLP}^{i,2},\cdots,\text{MLP}^{i,T^i}\}, \\ \text{MLP}^{i,j} = \mathbb{R}^{j} \to \mathbb{R},
    \mathbb{X}^i = \text{Causal-Mixing}^i(\tX^i).
\end{gather}
The Causal-Mixing is applied along the temporal dimension, which maps $X^i_{1:j}$ through $\text{MLP}^{i,j}$ to $\mathbb{X}^i_j$. The new representation $\mathbb{X} \in \mathbb{R}^{N \times T^i \times F}$ keeps the original shape.

Then their feature dimensions are mapped into the high-dimension hidden spaces to enrich their representation capabilities, facilitating the extraction of more complex hidden relationships:
\begin{align}
\label{equal-5}
    \mathcal{X}^i = \mathbb{X}^i \cdot W^i,
\end{align}
where $\mathcal{X}^i \in \mathbb{R}^{N \times T^i \times d}$, $d$ is the dimension of the latent space. $W^i \in \mathbb{R}^{F \times d}$, constructing a linear projection to enhance representational capability.

\subsection{Adaptive Spatial-Temporal Hypergraph Construction}
For the representation $\mathcal{X}^i \in \mathbb{R}^{N \times T^i \times d}$ of $i$-th scale, the hypergraph of $i$-th scale $\mathcal{X}^i$ can be constructed through a global-shared matrix $H\in \mathbb{R}^{N \times K}$, where $N$ represents the number of stocks, $K$ represents the number of industries, and $H_{l,m}=1$ denotes that the $l$-th stock belongs to the $m$-th industry. The $i$-th Spatial-Temporal hypergraph $\mathcal{G}^i$ is formulated as:
\begin{align}
    \mathcal{G}^i = (H,\mathcal{X}^i,E^i),
\end{align}
where $\mathcal{X}^i$ is the representation of $i$-th scale and works as the nodes of the $\mathcal{G}^i$, $E^i\in\mathbb{R}^{K\times T^i \times d}$ is the hyperedge constructed through the stock-industry relationship matrix $H$:
\begin{gather}
    \text{scores}^i =  \text{Linear}^i(\mathcal{X}^i)\in \mathbb{R}^{N \times 1}, \text{Linear}^i = \mathbb{R}^{T^i \times d} \to \mathbb{R},\\
    H^i = H \odot\text{scores}^i + (1-H)\odot (-\infty),\\\mathcal{H}^i = \text{SoftMax}(H^i,\text{dim}=0), E^i = (\mathcal{H}^i)^T \cdot \mathcal{X}^i,
\end{gather}
where the matrix $H$ is first processed with a learnable renovation to adapt to the current representation $\mathcal{X}^i$. $H^i$ is the intermediate result obtained through endowing the $H_{l,m}=1$ parts with learnable $\text{scores}^i$ and $H_{l,m}=0$ parts with $-\infty$ to ensure zeros in the $\text{SoftMax}$ step. Then the $\mathcal{H}^i \in \mathbb{R}^{N \times K}$ with Continuous Weights is obtained, and the hyperedges $E^i\in\mathbb{R}^{K \times T^i \times d}$ of industries are generated through information integration from all the related stocks. We obtain all the Spatial-Temporal hypergraphs $\{\mathcal{G}^1, \mathcal{G}^2, \cdots \mathcal{G}^S\}$ for each scale as above mentioned.

\subsection{Hyperedge-based Moving Aggregation}
\label{Hyperedge-Based Lead-Lag Relationships Extraction}
After delineating different time series scales with Spatial-Temporal hypergraphs $\{\mathcal{G}^1, \mathcal{G}^2, \cdots \mathcal{G}^S\}$, we further consider the lead-lag relationships between industries under each scale to analyze the complex correlations in financial markets---see Figure~\ref{The specific lead-lag interaction process within the lead-lag window.} in Appendix~\ref{Illustration Figures}. 

We construct a message passing mechanism on the hyperedges to capture the lead-lag relationships between industries. First, we employ a fixed-size sliding window as the range for examining the lead-lag relationships. For each hyperedge, through the sliding window, we can delineate different patches, which serve as the smallest units for constructing the lead-lag relationships:
\begin{gather}
    \mathcal{P}^i = \text{SlidingWindow}(E^i,\text{size}=k^i,\text{stride}=1),
\end{gather}
where $\mathcal{P}^i\in\mathbb{R}^{K \times (T^i - k^i +1) \times k^i \times d}$ denotes the patches of $E^i$, containing local information with the size of $k^i$. From the perspective of causality, considering the relationship between the entire window and its last point, we can construct the relationship matrix of the leading part and the lagging part. We then mine the lead-lag relationships inside each patch by considering the correlations between $\mathcal{P}^i_{before}$ and $\mathcal{P}^i_{last}$:
\begin{gather}
    \mathcal{P}^i_{before} = \text{Reshape}(\mathcal{P})^i \in \mathbb{R}^{(T^i - k^i +1)\times [K \times k^i]   \times d},\\
    \mathcal{P}^i_{last}  = \text{Reshape}(\mathcal{P})^i_{-1} \in \mathbb{R}^{ (T^i - k^i +1) \times [K \times 1] \times d}.
\end{gather}
Subsequently, by using the message passing method to propagate through the leading part, we obtain the feature fusion among hyperedges:
\begin{gather}
    \tilde{A}^i = \mathcal{P}^i_{before} \otimes \mathcal{P}^i_{last} \in \mathbb{R}^{(T^i - k^i +1) \times [K \times k^i] \times  K},
    A^i = SoftMax(\tilde{A}^i, \text{dim}=1),\\\tilde{E}^i = A^i \otimes \mathcal{P}^i_{before} \in \mathbb{R}^{ K \times (T^i - k^i +1) \times d},
    \hat{E}^i \in \mathbb{R}^{  K \times T^i \times d} = 
    \begin{cases}
   E^i, & t < k^i \\
{\tilde{E}^i}, & k^i \le t \le T^i\end{cases},
\end{gather}
where $\tilde{A}^i$ considers the correlations among the leading parts and lagging parts, $A^i$  probabilityizes $\tilde{A}^i$ to control weightsuming,  $\hat{E}^i\in\mathbb{R}^{K \times T^i \times d}$ denotes the embedding of hyperedges integrated with lead-lag relationships among different industries. It is particularly important to note that when $t < k^i$, no transformation is performed; when $k^i \le t \le T^i$, the result incorporating lead-lag interactions is used as the new hyperedges.

This is the method of considering the leading and lagging relationships between industries under each scale using Spatial-Temporal hypergraphs. After performing the above operation for each scale, we obtain a multi-scale representation of hyperedges $\{\hat{E}^1, \hat{E}^2,\cdots, \hat{E}^S\}$. Subsequently, fusion between scales will be carried out to integrate information of different granularities.

\subsection{Hyperedge-based Multi-scale Fusion}
\label{Hyperedge-Based Multi-Scale Fusion}
For multi-scale representation of hyperedges $\{\hat{E}^1, \hat{E}^2,\cdots, \hat{E}^S\}$ which contains the industry information, we further extract the potential correlations among different scales to reveal the latent interactions among different industries. First, we up-sample different scales back to the original length to ensure uniform representation size:
\begin{gather}
    \hat{\mathcal{E}}^i = \text{Linear}^i(\hat{E}^i) \in \mathbb{R}^{K \times T \times d}, \text{Linear}^i = \mathbb{R}^{T^i} \to \mathbb{R}^{T},\\
    \hat{\mathcal{E}} = \{\hat{\mathcal{E}}^1,\hat{\mathcal{E}}^2,\cdots,\hat{\mathcal{E}}^S\} \in \mathbb{R}^{S \times K \times T \times d}.
\end{gather}
Then, we construct adaptive adjacency matrices for these hyperedges of different scales to measure the correlations between different industries at various information granularities. We employ metric learning for this task and use a learnable Mahalanobis distance~\cite{goldberger2004neighbourhood,qiu2025duet} to automatically build relationships between different industries in the hidden space. Since the operations across the time dimension $T$ are independent and can be executed in parallel, we omit this dimension in the following formulations for notational simplicity.
\begin{gather}
    \tilde{\mathcal{E}} = \text{Flatten}(\hat{\mathcal{E}})\in\mathbb{R}^{(S \times K) \times d}, D_{ij} =  (\tilde{\mathcal{E}}_i - \tilde{\mathcal{E}}_j)\cdot Q \cdot(\tilde{\mathcal{E}}_i - \tilde{\mathcal{E}}_j)^T ,\\
     C_{ij} = \begin{cases}
    \frac{1}{D_{ij}} & \text{if } i \ne j \\
    0 & \text{if } i = j
    \end{cases},~~B = \text{SoftMax}(C,\text{dim}=0),
\end{gather}
where $D,C,B \in \mathbb{R}^{(S\times K)\times (S\times K)}$.  $Q\in \mathbb{R}^{d\times d}$ is a learnable semi-positive definite matrix, and can be practically constructed by $Q =  A^T \cdot A$, where A is also a learnable matrix. Subsequently, we construct interactions between hyperedges at different scales using message passing and further update the representation of hyperedges.
\begin{gather}
    % \tilde{B} = \mathcal{D}^{-1}\otimes B \otimes \mathcal{D}, 
    \mathcal{E}^\prime = B\cdot\tilde{\mathcal{E}}\cdot W^e\in \mathbb{R}^{(S \times K)\times d},
\end{gather}
where 
% $\mathcal{D} \in \mathbb{R}^{(S\times K)}$ is the degree matrix,
$W^e \in \mathbb{R}^{d\times d}$ is the learnable parameters in the graph message passing mechanism. After integrating the lead-lag relationships between industries on the hyperedges and fusing the hidden relationships between different scales, the representation of the hyperedges $\mathcal{E}^\prime$ is adaptively updated. We then continue to complete the message passing from hyperedges to nodes---see Figure~\ref{Three types of message passing} in Appendix~\ref{Illustration Figures}, thereby feeding back the complex correlations within the industry to specific stocks and completing the update of the representation of stock nodes. Finally, we restore the omitted $T$ dimension:
\begin{align}
\label{equal-20}
    \mathcal{V} &= \mathcal{E}^\prime \otimes \mathcal{H}\in \mathbb{R}^{(S\times N) \times T \times d}.
\end{align}

\subsection{Predictor}
\label{Forecasting Head}
% 在使用基于超图的结构完成了复杂的表征提取滞后，我们采用线性预测器对于未来的每个结构的金融指标进行预测：
After completing the complex representation extraction using a hypergraph-based structure, we employ an MLP-based predictor and incorporate residual connection modules to forecast financial indicators across each scale $i$.
\begin{gather}
    \hat{\mathcal{X}}^i =\text{MLP}^{i}_{1}(\mathcal{X}^i)\in\mathbb{R}^{N \times 1},\label{equal-27}\\\tilde{\mathcal{V}}^i = \text{MLP}^{i}_{2}(\mathcal{V}^i)\in\mathbb{R}^{N \times 1},\label{equal-28}\\
    \hat{Y}^i = \hat{\mathcal{X}}^i + \tilde{\mathcal{V}}^i,~~ \hat{Y} = \sum_{i=1}^{S} \hat{Y}^i,
\end{gather}
where $\mathcal{X}^i \in \mathbb{R}^{(N \times T^i \times d)}$ represents the initial node embedding in Equation~\ref{equal-5} that does not consider the lead-lag relationship and multi-scale interaction effects. However, $\mathcal{V}^i \in \mathbb{R}^{(N \times T \times d)}$ is the spatio-temporal hypernode embedding in Equation~\ref{equal-20} that incorporates lead-lag correlation and multi-scale information. Finally, we combine the prediction results at different scales to obtain the final prediction result.

\begin{table*}[!t]
\centering
\caption{Comparison results on public stock market datasets (measured by t-test with p-value < 0.01). The methods for comparison are mainly divided into five types: HGNN (HyperGraph Neural Network), Transformer, MLP (Multi-Layer Perceptron), GNN (Graph Neural Network), and RNN (Recurrent Neural Network). Bold indicates the best (SOTA) results.}
% \caption{Comparison results on public stock market datasets ($p < 0.01$ via t-test). The baselines include: \textbf{HGNN} (Hermes, ESTIMATE [EST.], STHAN-SR [STHAN.], ASHyper), \textbf{Transformer} (iTransformer [iTrans.], PatchTST, DUET), \textbf{MLP} (StockMixer [SMixer.], DLinear [DLin.], TimeMixer [TMixer.]), \textbf{GNN} (RSR-I, GAT, RGCN), and \textbf{RNN} (ALSTM, LSTM, SegRNN [SegR.]). Bold indicates the best performance.}
\resizebox{1\columnwidth}{!}{
\begin{tabular}{cc|cccc|ccc|ccc|ccc|cc}
\toprule 
\multicolumn{2}{c|}{\multirow{2}{*}{\textbf{Models}}}& \multicolumn{4}{c|}{\textbf{HGNN}} & \multicolumn{3}{c|}{\textbf{Transformer}} & \multicolumn{3}{c|}{\textbf{MLP}} & \multicolumn{3}{c|}{\textbf{GNN}} & \multicolumn{2}{c}{\textbf{RNN}} \\ 
\cmidrule{3-17} 
\multicolumn{2}{c|}{}  & \textbf{Hermes} & \textbf{EST.} & \textbf{STHAN.} & \textbf{ASHyper} & \textbf{iTrans.} & \textbf{PatchTST} & \textbf{DUET} & \textbf{SMixer.} & \textbf{DLin.} & \textbf{TMixer.} & \textbf{RSR-I} & \textbf{GAT} & \textbf{RGCN} & \textbf{ALSTM} & \textbf{LSTM} \\
\midrule
\multirow[c]{3}{*}{\rotatebox{90}{\textbf{NASDAQ}}}  & IC & \textbf{0.044} & 0.037 & 0.039 & 0.041 & 0.038 & 0.039 & 0.039 & 0.043 & 0.031 & 0.041 & 0.038 & 0.035 & 0.034 & 0.035 & 0.032 \\\cmidrule{3-17}
 & RIC & \textbf{0.538} & 0.444 & 0.451 & 0.508 & 0.420 & 0.496 & 0.494 & 0.501 & 0.453 & 0.532 & 0.398 & 0.377 & 0.382 & 0.371 & 0.354  \\\cmidrule{3-17}
& SR & \textbf{2.161} & 1.307 & 1.416 & 1.157 & 1.877 & 1.261 & 1.751 & 1.465 & 1.778 & 1.177 & 1.238 & 1.233 & 1.054 & 0.941 & 0.892 \\
\cmidrule{1-17}
\multirow[c]{4}{*}{\rotatebox{90}{\textbf{NYSE}}} & IC & \textbf{0.032} & 0.030 & 0.029 & 0.025 & 0.023 & 0.029 & 0.024 & 0.029 & 0.026 & 0.031 & 0.026 & 0.025 & 0.025 & 0.023 & 0.024  \\\cmidrule{3-17}
 & RIC & \textbf{0.466} & 0.327 & 0.344 & 0.250 & 0.303 & 0.284 & 0.327 & 0.351 & 0.277 & 0.314 & 0.284 & 0.297 & 0.275 & 0.276 & 0.256  \\\cmidrule{3-17}
 & SR & \textbf{1.655} & 1.115 & 1.228 & 1.208 & 1.520 & 0.898 & 1.361 & 1.454 & 1.526 & 1.556 & 0.098 & 1.070 & 0.932 & 0.764 & 0.857  \\
\cmidrule{1-17} 
\multirow[c]{3}{*}{\rotatebox{90}{\textbf{S\&P500}}} & IC & \textbf{0.050} & 0.035 & 0.037 & 0.046 & 0.048 & 0.046 & 0.049 & 0.041 & 0.043 & 0.040 & 0.033 & 0.034 & 0.028 & 0.029 & 0.031 \\\cmidrule{3-17}
 & RIC & \textbf{0.334} & 0.241 & 0.227 & 0.268 & 0.242 & 0.249 & 0.275 & 0.262 & 0.214 & 0.253 & 0.200 & 0.191 & 0.175 & 0.181 & 0.186  \\\cmidrule{3-17}
 & SR & \textbf{2.247} & 1.547 & 1.533 & 2.148 & 1.326 & 1.402 & 1.982 & 1.586 & 1.557 & 2.170 & 1.437 & 1.484 & 1.359 & 1.298 & 1.332 \\
\bottomrule
\addlinespace
\multicolumn{17}{l}{Model abbreviations: EST. (ESTIMATE), STHAN. (STHAN-SR), iTrans. (iTransformer), SMX (StockMixer), DLin. (DLinear), and TMX (TimeMixer).}
\end{tabular}}
\label{Comparison results on stock metrics}
\end{table*}

\subsection{Optimization Objectives}
\label{Optimization Objectives}
Following StockMixer~\cite{StockMixer}, We use the 1-day return ratio of a stock as the ground-truth rather than the normalized price used in previous work using a combination of a pointwise regression and pairwise ranking-aware loss to minimize the MSE between the predicted and actual return ratios while maintaining the relative order of top ranked stocks with higher expected return for investment as:
\begin{gather}
    L = L_\text{MSE} + \alpha  \displaystyle\sum_{i=1}^N\sum_{j=1}^N \max (0,-(\hat{r}_i^t - \hat{r}_j^t)(r_i^t - r_j^t)),
\end{gather}
where $\hat{r}^t$ and ${r}^t$ are the predicted and actual ranking scores, respectively, and $\alpha$ is a weighting parameter.

\section{Experiments}
\label{Experiments}

\subsection{Experimental Setup}
\label{Experimental Setup}

\noindent
\textbf{Datasets and Baselines.}
 We evaluate our approach using three real-world datasets from the US stock market. These datasets contain relatively complete stock-industry relationships or Wiki-based company relations. Besides, we conduct a comparative analysis between our proposed Hermes and a range of baseline methods, which include prominent approaches in both machine learning and deep learning methods for STSF. Given that existing research has demonstrated the limited effectiveness of traditional statistical methods~\cite{StockMixer,li2024master}, we exclude them from our comparative evaluation to focus on more competitive approaches.

\noindent
\textbf{Implementation Details.}
For fair comparison, all samples are generated by moving a 16 time steps lookback window. All experiments with Hermes are conducted using PyTorch~\cite{paszke2019pytorch} in Python 3.9, and are executed on a server featuring an Intel(R) Xeon(R) Platinum 8358 CPU and an NVIDIA Tesla-A800 GPU. We use ADAM~\cite{Adam} with an initial learning rate of 5e-3. Detailed descriptions of the experimental datasets, baselines, and evaluation metrics are provided in the Appendix~\ref{Experimental Setup}.

\begin{table*}[t]
\centering
\caption{Ablation study results on public datasets NASDAQ, NYSE, and S\&P500.}
\resizebox{0.8\columnwidth}{!}{
\begin{tabular}{c|ccc|ccc|ccc}
\toprule
\textbf{Ablation} & \multicolumn{3}{c|}{\textbf{NASDAQ}} & \multicolumn{3}{c|}{\textbf{NYSE}}  & \multicolumn{3}{c}{\textbf{S\&P500}}\\  \midrule
\textbf{Model Component} & \textbf{IC} & \textbf{ICIR}  &\textbf{SR} & \textbf{IC} & \textbf{ICIR}  & \textbf{SR} & \textbf{IC} & \textbf{ICIR}  & \textbf{SR}\\
\midrule
% LSTM & 0.032 & 0.354 & 0.892& 0.024 & 0.256&0.857 &0.031 &0.186 &1.332\\
\textbf{w/o Multi-Scale Fusion} &0.038&0.464 &1.374&0.028&0.421&1.354&0.031&0.248&1.517  \\
\textbf{w/o Total Multi-Scale} & 0.030&0.422&0.975&0.023&0.347&1.199&0.032&0.213&1.075\\
\textbf{w/o Lead-Lag} & 0.039 & 0.472 &1.451& 0.028 &0.420&1.225&0.037&0.201&1.628\\
\textbf{w/o Skip-Connection} & 0.041&0.441&1.655&0.029&0.345&1.449&0.043&0.252&1.653 \\ 
\textbf{Only Initial Node Embedding} & 0.019&0.174&0.695&0.016&0.113&0.449&0.018&0.122&0.659 \\
\midrule
% \textbf{STHAN-SR} & 0.039 & 0.451 &1.416& 0.029 & 0.344 &1.228&0.037 &0.227&1.533\\
% \textbf{ESTIMATE} & 0.037 & 0.444 &1.307 & 0.030 &0.327&1.115 &0.035&0.241& 1.547\\
\textbf{Hermes (Ours)} & \textbf{0.044} & \textbf{0.538} & \textbf{2.161} & \textbf{0.032} & \textbf{0.466} & \textbf{1.655} & \textbf{0.050} & \textbf{0.334}& \textbf{2.247}\\
\bottomrule
\end{tabular}
\label{table:ablation}}
\label{Ablation study results on NASDAQ and NYSE}
\end{table*}

\begin{table*}[t]
\centering
\caption{Efficiency analysis on public stock datasets, concerning the training time, inference time (measured in milliseconds per batch under the condition of a batch size being 1), and memory usage (quantified in GB) of different HGNN-based (HyperGraph Neural Network) models. Model abbreviations: EST. (ESTIMATE) and STHAN. (STHAN-SR).}
\resizebox{1\columnwidth}{!}{
\begin{tabular}{c|cccc|cccc|cccc}
\toprule
\textbf{Dataset} & \multicolumn{4}{c|}{\textbf{NASDAQ}} & \multicolumn{4}{c|}{\textbf{NYSE}}  & \multicolumn{4}{c}{\textbf{S\&P500}}\\\midrule
\textbf{Property} & \textbf{STHAN.} & \textbf{EST.} & \textbf{ASHyper} & \textbf{Hermes} &  \textbf{STHAN.} & \textbf{EST.} &\textbf{ASHyper} & \textbf{Hermes} & \textbf{STHAN.} & \textbf{EST.} & \textbf{ASHyper} & \textbf{Hermes} \\
\midrule
\textbf{Training Time} &257 ms&1,267 ms&288 ms &215 ms &384 ms&1,462 ms&378 ms&282 ms&297 ms&1,317 ms&314 ms&211 ms  \\
\textbf{Inference Time} & 52 ms&163 ms&58 ms&42 ms&73 ms&241 ms&71 ms&61 ms&32 ms&109 ms&36 ms&25 ms\\
\textbf{Memory Usage} & 2.15 GB & 9.62 GB &1.99 GB &0.98 GB &2.82 GB &10.25 GB&2.17 GB &1.14 GB &1.23 GB&4.52 GB&1.51 GB &0.54 GB\\
\bottomrule
\end{tabular}}
\label{Efficiency Analysis table5}
\end{table*}

\subsection{Overall Performance}
\label{Overall Performance}
% We present a detailed comparison of our approach (Hermes) with five types of baseline methods (HGNN, Transformer, MLP, GNN, and RNN) in Table~\ref{Comparison results on stock metrics}. 

We conduct a comprehensive comparison of Hermes against five categories of baseline methods (HGNN, Transformer, MLP, GNN, and RNN) across three representative stock market datasets: NASDAQ, NYSE, and S\&P 500. As demonstrated in Table~\ref{Comparison results on stock metrics}. We have the following observations: 1) Hermes consistently achieves State-of-the-Art (SOTA) performance across all datasets and evaluation metrics. 2) For univariate methods, whether it is LSTM or ALSTM, their performance is inferior to that of other hybrid architectures. This strongly highlights the necessity and importance of capturing the complex industry and stock relationships in financial markets. 3) For GNNs algorithms such as RGCN, GAT, and RSR-I, their performance shows a significant improvement over RNNs, indicating that GNNs have strong stock relationship modeling capabilities. However, compared to HGNNs algorithms like STHAN-SR, ESTIMATE, and ASHyper, GNNs perform slightly worse. This suggests that in the context of the financial market industry, using hypergraph models to unify the relationships of stocks within the industry plays a crucial role in enhancing forecasting performance. 4) Compared to other hypergraph algorithms that consider stock relationships within industries (such as STHAN-SR, ESTIMATE, and ASHyper), the Hermes algorithm we propose shows significant performance improvement. This clearly demonstrates that, based on the consideration of stock relationships within industries, incorporating inter-industry lead-lag relationships and multi-scale relationships plays a crucial role in enhancing forecasting performance.

\subsection{Ablation Studies}
\label{Ablation Study}
Table~\ref{Ablation study results on NASDAQ and NYSE} illustrates the unique impact of each module. We have the following observations: 1) When the \textbf{multi-scale fusion module} in Section~\ref{Hyperedge-Based Multi-Scale Fusion} and the \textbf{total multi-scale module} (Remove the Total Multi-Scale module, including Decomposition in Section~\ref{Multi-Scale Decomposition} and Fusion in Section~\ref{Hyperedge-Based Multi-Scale Fusion}, respectively) are removed, the performance of the model decreases in both cases. This underscores the effectiveness of our multi-scale approach, confirming that both multi-scale decomposition and fusion are essential to the model's predictive power. 2) The ablation of the \textbf{lead–lag module} in Section~\ref{Hyperedge-Based Lead-Lag Relationships Extraction} causes a substantial decline in performance, demonstrating that explicitly modeling inter-industry lead–lag relationships is essential for strong predictive accuracy. 3) Removing the \textbf{skip-connection module} in Section~\ref{Forecasting Head} also reduces performance, confirming its crucial role in facilitating information transmission and gradient flow, thereby improving the model's feature extraction capabilities. 4) When the Predictor relies \textbf{solely on the initial node embedding} from Equation~\ref{equal-27} without integrating the updated features from Equation~\ref{equal-28}, the model performance undergoes a significant degradation. This is expected given the inherent complexity of stock forecasting, where primitive encodings are insufficient to capture intricate market dynamics. Such a sharp decline underscores the necessity and effectiveness of our proposed hypergraph framework.

\subsection{Efficiency Analysis}

\label{Efficiency Analysis}
In order to ensure good applicability of the model in practical applications, the runtime and memory usage of the model are crucial. We conducted an efficiency comparison analysis of Hermes and other hypergraph-based methods, STHAN-SR, ESTIMATE, and ASHyper---see Table~\ref{Efficiency Analysis table5}.

\textbf{Runtime Analysis:} In practical application scenarios, fast response is one of the key factors. For example, in the field of financial market forecasting, timely obtaining forecasting results can provide valuable decision-making basis for investors. If the model takes too long to operate, even if the prediction accuracy is high, it will be difficult to meet the actual needs. Therefore, we evaluate the computational speed by recording the time each model takes to train and infer a sample (that is, the batch size is uniformly set to 1) with the same dataset. The results show that the Hermes model performs excellently. Compared with the other two hypergraph-based methods, its computational time is reduced. This is mainly attributed to the fact that some modules of the Hermes model adopt an efficient MLP structure in the algorithm design process. This structure can significantly improve the computational efficiency while ensuring the accuracy of forecasting results.

\textbf{Memory Analysis:} Memory usage is also an important factor to measure the efficiency of the model. Excessive memory usage will not only limit the application range of the model in different hardware environments but also may cause the system to run slowly or even crash. In this experiment, we monitored the memory usage of each model in detail during operation. The results show that the Hermes has a lower memory usage rate. This advantage primarily stems from its unique architectural design, which enables superior performance without requiring large hidden dimensions, as further validated in Section~\ref{Hyperparameter Sensitivity} through sensitivity analysis of the Latent Space Dimension parameter. Consequently, Hermes effectively reduces memory usage while maintaining high performance, enhancing both the practicality and deployment flexibility of the model.

\section{Conclusions}
\label{Conclusions}
Stock time series forecasting plays a pivotal role in the modern economy, serving as a key tool for investors, financial institutions, and policymakers to make accurate decisions. However, existing forecasting methods often fall short in fully considering the lead-lag between industries and multi-scale information, resulting in limited predictive performance. To address these challenges, this study introduces the Hermes framework. By ingeniously fusing multi-scale analysis with lead-lag modules within a hypergraph network, the Hermes framework significantly boosts the performance of stock forecasting. Specifically, the hyperedge-based moving aggregation module incorporates a sliding window and utilizes dynamic temporal aggregation operations to consider lead-lag dependencies among industries. Meanwhile, the hyperedge-based multi-scale fusion module decomposes raw data at multiple scales, and then employs cross-scale, edge-to-edge message passing to integrate information from different scales while preserving the consistency of each scale.  

\clearpage


\bibliography{reference}
\bibliographystyle{unsrtnat}

\clearpage
\appendix
\section{Appendix}

\subsection{Experimental Setup}
\label{Experimental Setup}
\subsubsection{Datasets}
To validate the performance of Hermes, we use three publicly available stock market datasets, with the dataset statistics provided in Table~\ref{Statistics of datasets1}. These datasets provide a diverse set of stock time series, allowing us to comprehensively assess the model's effectiveness across different financial scenarios.
We evaluate our approach using three real-world datasets from the US stock market. These datasets contain relatively complete stock-industry relationships or Wiki-based company relations. The NASDAQ and NYSE datasets~\cite{feng2019temporal} include EOD data from 01/02/2013 to 12/08/2017, filtered from the respective markets. Abnormal patterns and penny stocks were removed, while preserving their representative properties, with NASDAQ being more volatile and NYSE more stable. The S\&P500 dataset~\cite{huynh2023efficient} gathers historical price data and industry information for the companies listed in the S\&P500 index from the Yahoo Finance database, covering the period from 01/04/2016 to 05/25/2022.

\begin{table}[h]
\caption{Statistics of datasets.}
\label{Statistics of datasets1}
\centering
\resizebox{0.55\columnwidth}{!}{
\begin{tabular}{c|c|c|c}
\toprule
\diagbox{\textbf{Property}}{\textbf{Dataset}} & \textbf{NASDAQ} & \textbf{NYSE} & \textbf{S\&P500}   \\
\midrule
\makecell{\# Stocks } & 1026 & 1737 & 474 \\ \midrule
\# Industries & 113 & 130 & 11 \\\midrule
\# Technical Indicators & 5 & 5 & 5\\\midrule
Start Time & 13-01-02 & 13-01-02 & 16-01-04 \\\midrule
End Time & 17-12-08 & 17-12-08 & 22-05-25\\\midrule
Total Time Steps & 1281 & 1281 & 1611 \\ \midrule
Train Time Steps & 756 & 756 & 1006 \\\midrule
Validate Time Steps & 252 & 252 & 253 \\\midrule
Test Time Steps & 273 & 273 & 352 \\
\bottomrule
\end{tabular}}
\end{table}

\subsubsection{Metrics}
Previous studies use different evaluation metrics, making it challenging to perform a comprehensive comparison of various methods. To provide a thorough assessment of the methods performance, we use four of the most commonly applied and reliable metrics: \textbf{IC} and \textbf{ICIR}, which are rank-based evaluation metrics; and \textbf{SR}, which is return-based.
% Previous studies used different evaluation metrics, making it challenging to perform a comprehensive comparison of various methods. To provide a thorough assessment of the techniques' performance, we use three of the most commonly applied and reliable metrics: \textbf{IC} and \textbf{ICIR}, which are rank-based evaluation metrics; and \textbf{SR}, which is return-based.
\begin{itemize}[left=0.1cm]
\item \textbf{Information Coefficient (IC):} IC is a coefficient that shows how close the prediction is to the actual result, computed by the average pearson correlation coefficient.
\begin{align}
IC_t = \frac{1}{N}\frac{(\hat{Y}_t - mean(\hat{Y}_t))^T(Y_t - mean(Y_t))}{std(\hat{Y}_t) \cdot std(Y_t)},
\end{align}
where $Y_t$ are the raw stock price trends and $\hat{Y}_t$ are the model predictions at each timestamp. We report the average IC over all test dataset.

\item \textbf{Information Ratio-based Information Coefficient (ICIR):} The information ratio of the IC metric, calculated by: 
\begin{align}
ICIR = \frac{mean(IC)}{std (IC)}.
\end{align}

\item \textbf{Sharpe Ratio (SR):} Sharpe ratio~\cite{sharpe1994sharpe} measures the profitability of the investment method and take into account risk.
\begin{align}
SR = \frac{E_t[\rho_t - \rho_F]}{\sqrt{\mathrm{var}_t(\rho_t - \rho_F)}}, 
\end{align}
where $\rho_t = \frac{p_t}{p_{t-1}} - 1$ is the return on the portfolio and $p_F$ is the return on the risk-free asset which is always 0.
\end{itemize}

\subsubsection{Baselines}
We conduct a comparative analysis between our proposed Hermes and a range of baseline methods, which include prominent approaches in both machine learning and deep learning methods for STSF. Given that existing research has demonstrated the limited effectiveness of traditional statistical methods~\cite{StockMixer,li2024master}, we exclude them from our comparative evaluation to focus on more competitive approaches.

\begin{itemize}[left=0.1cm]
\item \textbf{LSTM}~\cite{hochreiter1997long} refers to the standard LSTM model, which operates on sequential data, including closing prices and moving averages over 5, 10, 20, and 30 days, to generate a sequential embedding. A fully connected (FC) layer is then employed to predict the return ratio.
\item \textbf{ALSTM}~\cite{DBLP:conf/ijcai/FengC0DSC19} integrates adversarial training and stochastic simulation into an enhanced LSTM model, allowing it to more effectively capture market dynamics.
\item \textbf{RGCN}~\cite{li2021modeling} introduces a model that captures both positive and negative correlations among stocks using a correlation matrix computed from historical market data. The LSTM mechanism in RGCN layers helps mitigate over-smoothing issues when predicting overnight stock price movements. 
\item \textbf{GAT}~\cite{velivckovic2017graph} employs Graph Attention Networks (GAT) to aggregate stock embeddings, which are encoded by a GRU, on the stock graph. By combining the advantages of Graph Neural Networks (GNN) and attention mechanisms, it enhances performance in large-scale graph settings.
\item \textbf{RSR}~\cite{feng2019temporal} introduces a novel neural network-based framework called Relational Stock Ranking, which addresses the stock forecasting problem in a learning-to-rank approach. Additionally, it proposes a new component in neural network modeling, called Temporal Graph Convolution, which is designed to explicitly capture the domain knowledge of stock relationships in a time-sensitive manner.
\item \textbf{STHAN-SR}~\cite{sawhney2021stock} introduces a novel Spatio-Temporal Hypergraph Attention Network that models inter-stock relationships of different types and strengths as a hypergraph for stock ranking. It combines temporal Hawkes attention with spatial hypergraph convolutions through hypergraph attention to capture both the correlations in the movements of related stocks and the temporal evolution of their historical features.
\item \textbf{ESTIMATE}~\cite{huynh2023efficient} integrates temporal generative filters within a memory-based shared parameter LSTM network, enhancing the model's ability to learn temporal patterns for each stock. Furthermore, it introduces attention hypergraph convolutional layers that utilize the wavelet basis—a convolution approach that leverages the polynomial wavelet basis to streamline message passing and focus on localized convolutions.
\item \textbf{Linear}~\cite{zeng2023transformers} employs a straightforward approach by utilizing only fully connected layers to predict the final price. This method does not incorporate any advanced architectures or temporal dependencies, relying solely on basic linear transformations to generate the forecasting. Despite its simplicity, it serves as a baseline model for comparison with more complex models.
\item \textbf{StockMixer}~\cite{StockMixer} introduces a lightweight yet effective MLP-based architecture for stock price forecasting. The model incorporates three key components—indicator mixing, time mixing, and stock mixing—which work together to capture the intricate correlations within the stock data.
\item \textbf{DLinear}~\cite{zeng2023transformers} employs a straightforward linear regression approach that challenges complex Transformer-based architectures. It decomposes the time series and directly regresses historical data for future prediction using a weighted sum operation, demonstrating that simple linear formulations can effectively capture long-term temporal trends.
\item \textbf{iTransformer}~\cite{DBLP:conf/iclr/LiuHZWWML24} applies attention and feed-forward networks on inverted dimensions to better capture multivariate correlations. Specifically, it embeds the time series of each variable into variate tokens to model dependencies among variables via attention, while utilizing feed-forward networks to learn nonlinear representations for each token.
\item \textbf{TimeMixer}~\cite{DBLP:conf/iclr/WangWSHLMZ024} utilizes a fully MLP-based architecture designed to leverage disentangled multiscale time series. It incorporates Past-Decomposable-Mixing (PDM) blocks to mix seasonal and trend components across different scales, and Future-Multipredictor-Mixing (FMM) blocks to aggregate predictions, thereby capturing both microscopic and macroscopic information effectively.
\item \textbf{PatchTST}~\cite{DBLP:conf/iclr/NieNSK23} presents an efficient Transformer-based design centered on patching and channel independence. It segments time series into subseries-level patches to serve as input tokens, allowing the model to attend to longer histories with reduced computational costs while retaining local semantic information through independent channel processing.
\item \textbf{ASHyper}~\cite{DBLP:conf/nips/ShangCWC24} proposes an Adaptive Multi-Scale Hypergraph Transformer to model group-wise interactions in time series. It features an adaptive hypergraph learning module and a multi-scale interaction module to capture comprehensive patterns. Additionally, it employs node and hyperedge constraints to cluster nodes with similar semantics and differentiate temporal variations.
\item \textbf{DUET}~\cite{qiu2025duet} introduces a dual clustering framework that operates on both temporal and channel dimensions to address pattern heterogeneity. It employs a Temporal Clustering Module (TCM) to group time series into fine-grained distributions and a Channel Clustering Module (CCM) to capture inter-channel relationships in the frequency domain, effectively mitigating noise and modeling complex dependencies.
\end{itemize}

\begin{figure*}[h]
  \centering
  \subfloat[Lookback length]
  {\includegraphics[width=0.243\textwidth]{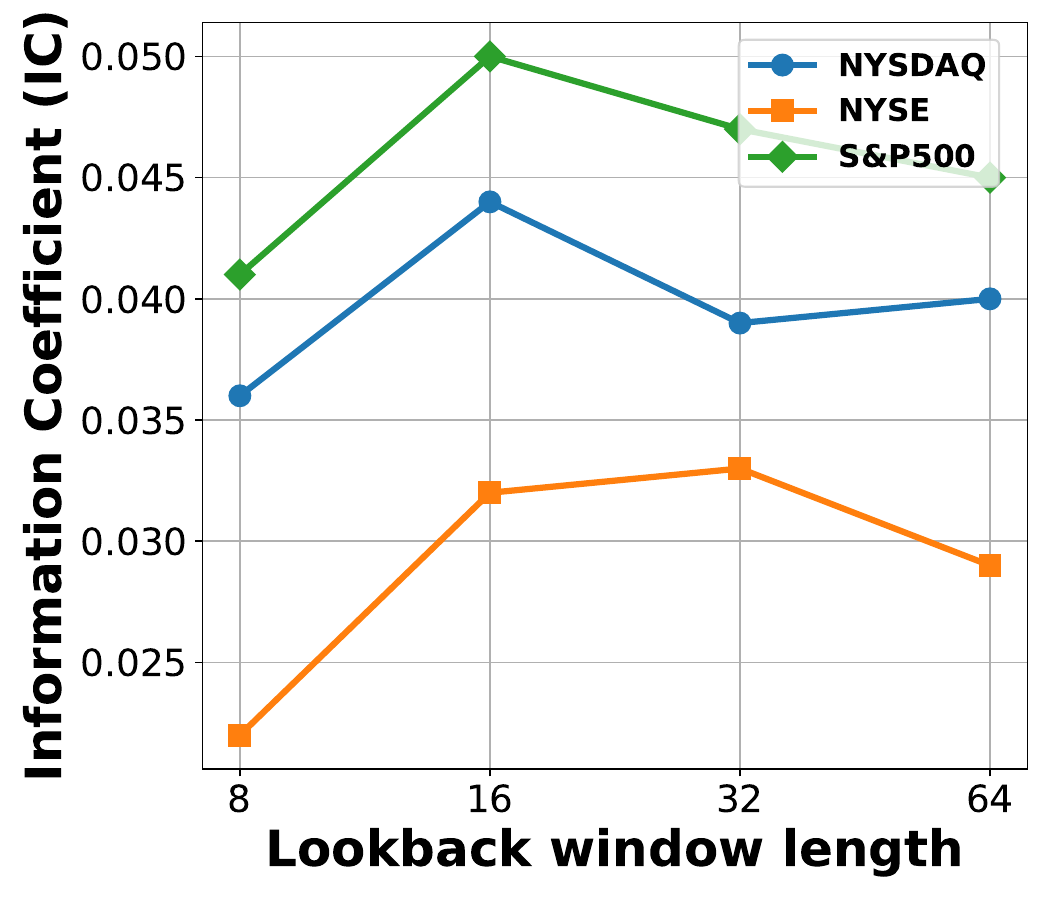}\label{Lookback window length}}
  \hspace{0.5mm}\subfloat[Latent space dimension]
  {\includegraphics[width=0.243\textwidth]{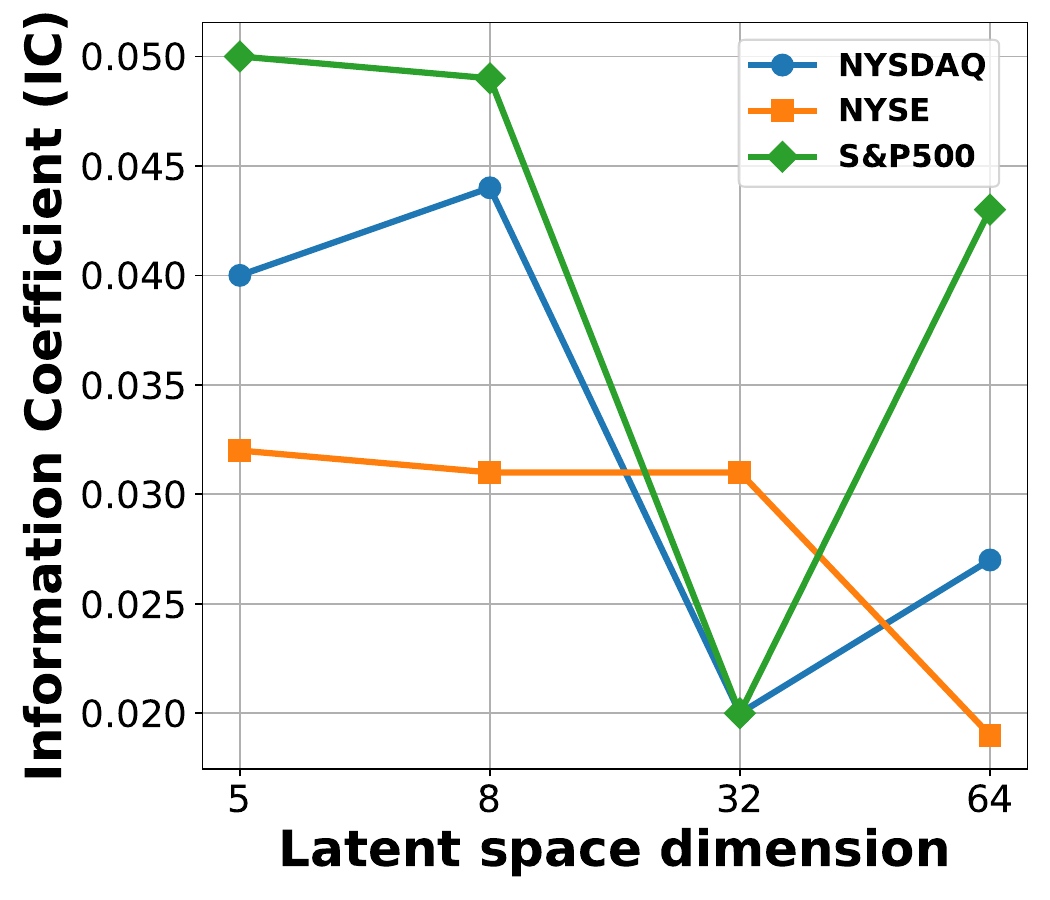}\label{Latent space dimension}}
  \hspace{0.5mm}\subfloat[Lead-Lag steps]
  {\includegraphics[width=0.243\textwidth]{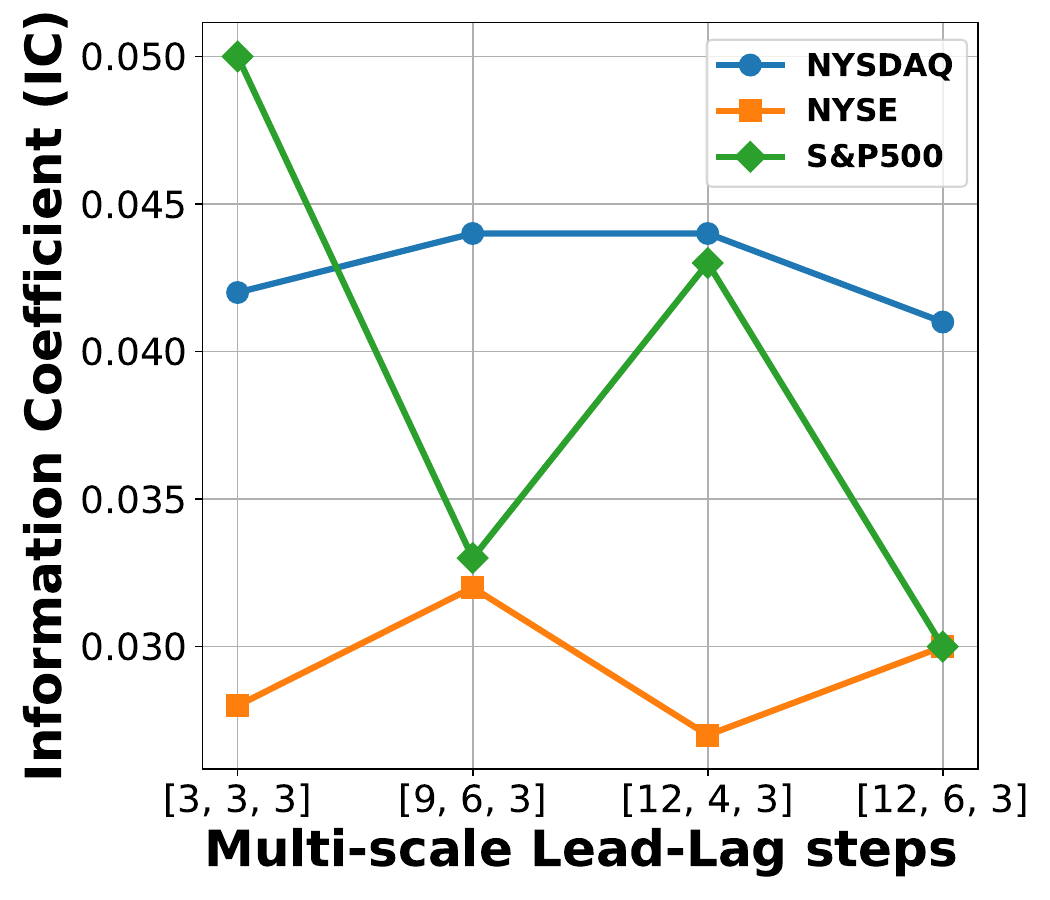}\label{Lead-Lag steps}}
 \hspace{0.5mm}\subfloat[Score weight]
  {\includegraphics[width=0.243\textwidth]{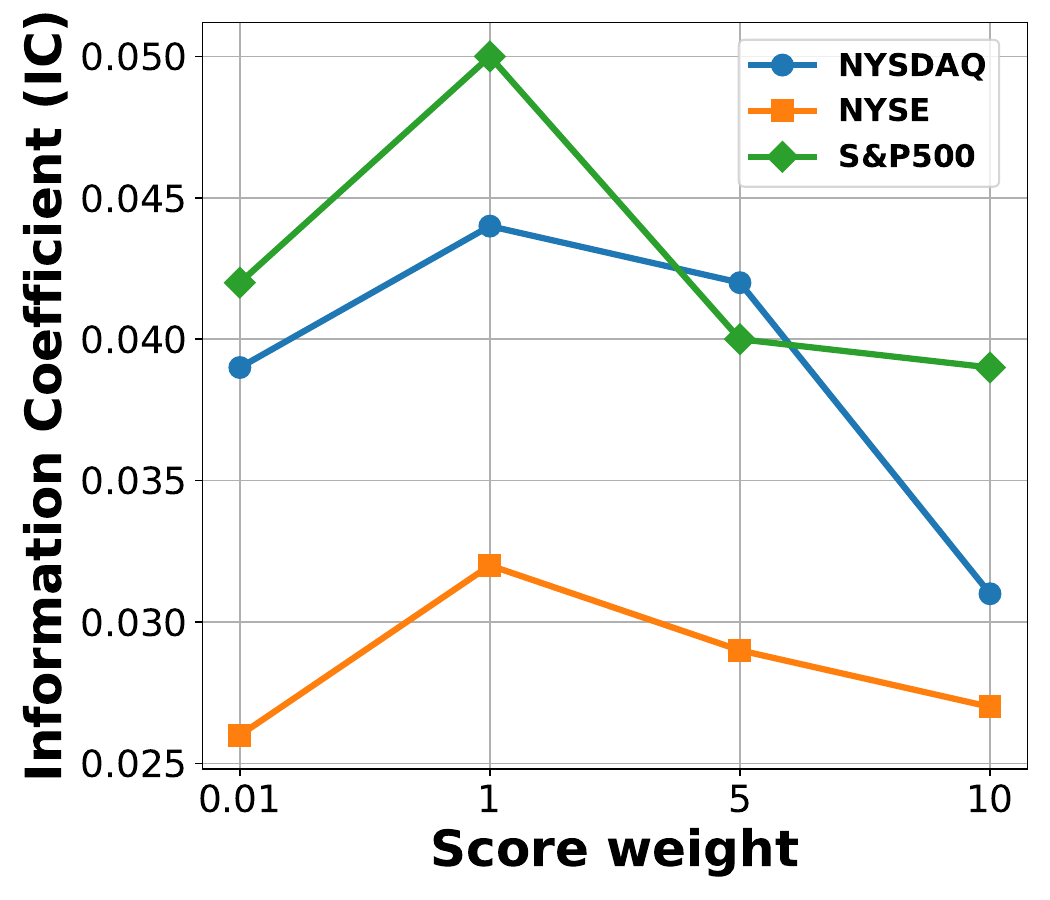}\label{Score weight}}
  \caption{Parameter sensitivity studies of main hyper-parameters in Hermes.}
    \label{Hyperparameter.}
\end{figure*}

\section{Limitations}
\label{app:limitations}

Despite the strong performance of Hermes on multiple real-world stock forecasting benchmarks, several limitations remain. Hermes relies on predefined industry relationships to construct hyperedges, which may not fully capture rapidly evolving or latent market structures in real financial systems. And, although the proposed multi-scale fusion and lead-lag aggregation modules improve forecasting accuracy, the model may still face scalability challenges when applied to extremely large-scale financial markets with thousands of assets and dense inter-industry interactions.

\section{Impact Statement}
\label{app:impact}

This work proposes Hermes, a spatial-temporal hypergraph framework for stock time series forecasting that explicitly models inter-industry lead-lag relationships and multi-scale market dynamics. The proposed framework may benefit financial forecasting systems, quantitative investment research, and market analysis by improving the modeling of complex stock correlations. This research is conducted for academic purposes only. All datasets used in this work are publicly available, and no private or sensitive personal information is involved. And we only use LLMs for grammatical polishing during writing, with all research content being original and ethically compliant.

\section{Related Work}
\label{Related Works}
% In this section, we first provide an overview of traditional stock forecasting methods, followed by a detailed discussion of the application of deep learning techniques in the field of stock forecasting. Subsequently, we summarize the latest advancements in deep learning for handling lead-lag relationships and constructing multi-scale models. Finally, we conduct a comparative analysis of other stock forecasting methods based on hypergraph theory.
\subsection{Traditional Stock Time Series Forecasting}
Early stock forecasting methods rely mainly on statistical learning techniques to capture patterns and relationships in time series~\cite{ye2016novel,ARIMA,markov,elton2009modern}. One of the most widely used techniques is the Autoregressive Integrated Moving Average (ARIMA)~\cite{ARIMA} model, which stabilizes data by combining Autoregressive components, Moving Average, and differencing, thereby modeling time series. Another commonly used traditional method is the Hidden Markov Model (HMM)~\cite{markov}, which is used to describe unobservable states in the market and can detect changes in market conditions, making it suitable for analyzing market transitions. Although these methods provide predictive capabilities, they exhibit significant limitations when dealing with nonlinear, dynamic changes, and sudden events in financial markets. As a market environment becomes more complex, traditional statistical methods increasingly fail to provide sufficient predictive accuracy and robustness. 

With the rapid development in machine learning, machine learning methods for stock time series forecasting have emerged~\cite{alkhatib2013stock,nugroho2014decision,kim2003financial}. For example, Support Vector Machines (SVMs)~\cite{kim2003financial} map data to high-dimensional spaces using kernel functions and find optimal separating hyperplanes in those spaces, which allows them to capture complex relationships and dynamics in financial markets. However, SVMs are highly sensitive to parameter selection and often struggles to capture long-term dependencies. Overall, although machine learning methods show considerable potential in stock forecasting, they still rely on manual feature engineering and model design. Moreover, these traditional stock forecasting methods have not delved into the exploration of lead-lag relationships and multi-scale modeling.

\subsection{Deep Learning Stock Time Series Forecasting}
% While deep learning has made impressive advances in NLP and CV~\cite{zhang2025imdprompter,zhang2024can,zhang2025rethinking,chen2024gim,zhang2024distilling}, demonstrating that learned features can outperform human-designed features, they also evolve over time series. 
With the rise of deep learning, notable progress has occurred in stock forecasting. This is mainly because deep neural networks can model complex nonlinear patterns. Some studies use recurrent neural networks (RNNs)~\cite{nelson2017stock,qin2017dual,akita2016deep,rahman2019predicting}  to capture the historical development in individual stock prices and predict their short-term trends. For example, Long Short-Term Memory (LSTM) networks combined with historical stock data and text information are used to predict stock prices~\cite{akita2016deep}. In addition, due to the significant progress of Multi-Layer Perceptrons (MLPs) architecture in the general time series field, the DLinear~\cite{zeng2023transformers}, CycleNet~\cite{lin2025cyclenet}, and SparseTSF~\cite{SparseTSF} show the benefits of this simple architecture for time series forecasting. Increasingly many proposals target stock forecasting performance by improving the MLPs architecture. For example, a series of studies~\cite{StockMixer,lazcano2024back,tashakkori2024forecasting} enhance the learning ability of simple MLPs structures by utilizing the MLP-Mixer backbone, thereby improving forecasting accuracy. Next, the Transformer model has become increasingly popular due to the ability of its self-attention mechanism to capture long-term dependencies. For example, Master~\cite{li2024master} proposes a novel stock transformer for stock price forecasting to effectively capture stock correlation.

Some deep learning models have begun to explore lead-lag relationships and multi-scale modeling. For instance, StockMixer~\cite{StockMixer} engages in multi-scale modeling by segmenting the original time series into subsequence-level patches and mixing features at different scales. However, this approach lacks effective integration of information across scales; it merely concatenates information from different scales without thorough fusion. On the other hand, ESTIMATE~\cite{huynh2023efficient} takes into account lead-lag relationships through data-driven detection of leading and lagging time series clusters. Nonetheless, this method falls short in capturing fine-grained lead-lag dynamics, as clustering can only reveal global lead-lag relationships. Distinctive from the aforementioned approaches, Hermes employs a sliding window technique and utilizes dynamic temporal aggregation operations to consider fine-grained lead-lag dependencies among stock industries. Moreover, it features a multi-scale fusion module, which utilizes a cross-scale, edge-to-edge message-passing to integrate information from different scales while maintaining the consistency of each scale.

While these deep learning methods have advanced stock forecasting, the complexity of stock data and the need to account for intricate relationships among stocks have led researchers to explore new methods based on hypergraphs.

\subsection{Hypergraph-based Stock Time Series Forecasting}
Traditional graphs have limitations when it comes to modeling higher-order multivariate relationships, as they can only represent relations between pairs of nodes. To address this issue, Hypergraph Neural Networks (HGNNs)~\cite{feng2019hypergraph} were introduced. Hypergraphs extend the concept of simple graphs to enable the capture of relationships among multiple stocks~\cite{chen2020hypergraph}. Hypergraphs have gained increasing attention and application across various fields. Hyperedge in a hypergraph represents a set of vertices, making them suitable for modeling non-pairwise relationships~\cite{luo2014stock,sawhney2020spatiotemporal}. For example, HGAM~\cite{li2022hypergraph} is a hypergraph-based reinforcement learning method for stock portfolio selection. THGNN~\cite{xiang2022temporal} is a temporal and heterogeneous graph neural network model that aims to learn dynamic relationships by using two-stage attention mechanisms. STHAN-SR~\cite{sawhney2021stock} enhances corporate relevance based on Wiki data and uses hypergraph convolution to propagate information from higher-order neighbors. DHSTN~\cite{liao2024stock} is a dynamic hypergraph network for learning spatio-temporal relations of stocks. Dynamic hypergraphs are generated adaptively using (Graph Attention Networks) GATs to capture time-varying higher-order spatial relationships among stocks. The latest method, ESTIMATE~\cite{huynh2023efficient}, employs hypergraphs to capture non-pairwise correlations, utilizing temporal generative filters to identify individual patterns for each stock. 

Unlike the above methods, we propose a new hypergraph-based stock forecasting model that specifically takes into account the multi-scale nature of stock data, enabling it to capture the complexity of market dynamics more comprehensively. Furthermore, our approach places particular emphasis on the lead-lag relationships between financial industries, where the changes in one industry may precede or lag behind those in another. By capturing these key relationships more fully, we aim to improve forecasting accuracy.

\subsection{Illustration Figures}
\label{Illustration Figures}
\begin{figure}[h]
    \centering
    \begin{minipage}[c]{0.48\textwidth}
        \centering
        \includegraphics[width=\linewidth]{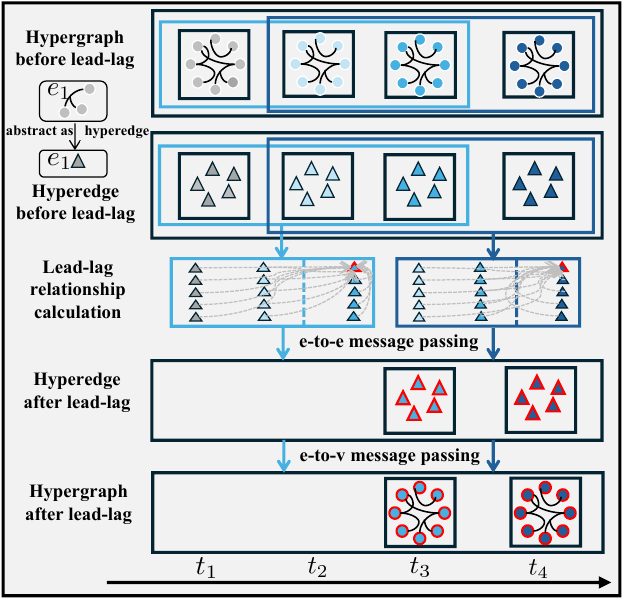}
        \caption{The specific lead-lag interaction process within the window, with a lead-lag step of 3 as an example.}
        \label{The specific lead-lag interaction process within the lead-lag window.}
    \end{minipage}
    \hfill % 水平间距
    \begin{minipage}[c]{0.48\textwidth}
        \centering
        \includegraphics[width=\linewidth]{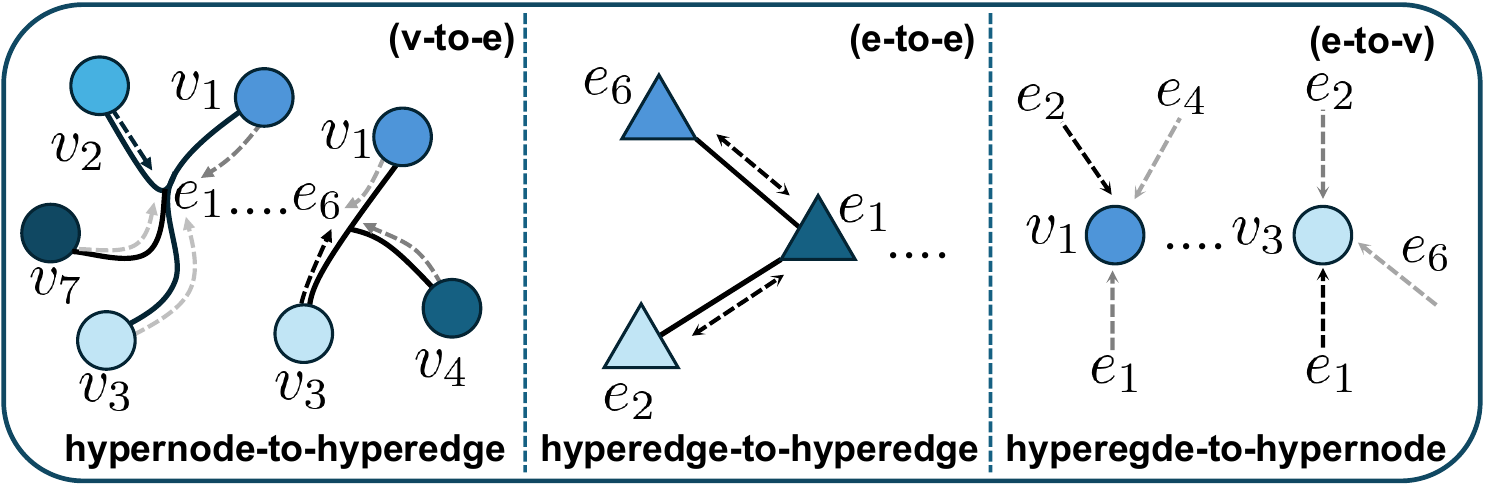}
        \caption{Three types of message passing: hypernodes for stocks and hyperedges for industries.}
        \label{Three types of message passing}
    \end{minipage}
\end{figure}

Figure~\ref{The specific lead-lag interaction process within the lead-lag window.} illustrates the specific lead-lag interaction process within the window, with a lead-lag step of 3 as an example. Figure~\ref{Three types of message passing} depicts three types of message passing: hypernodes for stocks and hyperedges for industries.

\subsection{Hyperparameter Sensitivity}
\label{Hyperparameter Sensitivity}
 We also study the parameter sensitivity of the Hermes---see Figure~\ref{Hyperparameter.}.

\textbf{Lookback window length $T$:} We examine the predictive performance of Hermes as we adjust the lookback window length, $T$, illustrated in Figure~\ref{Lookback window length}. Consistently across all datasets, a balanced window length emerges as the most effective. Window lengths that are too short result in a swift decline in performance due to inadequate information, whereas excessively lengthy sequences also underperform because of the absence of timely information benefits and the heightened learning expenses for stocks.

\textbf{Latent space dimension $d$:} In Section~\ref{Multi-Scale Decomposition}, we enhance the representational capacity of indicators by mapping them into a high-dimensional hidden space $d$, thereby facilitating the extraction of more intricate hidden relationships. In Figure~\ref{Latent space dimension}, we explore the impact of different hidden dimensions $d$ and observe that datasets exhibit optimal performance at varying $m$ values. Notably, when $d$ is set to smaller values such as 5 or 8, models tend to achieve good performance; however, as $d$ increases, models become prone to overfitting, leading to a decline in performance. Furthermore, selecting smaller $d$ not only improves model performance but also effectively reduces the number of model parameters.

\textbf{Multi-scale Lead-Lag steps $w^1,w^2,\cdots, w^S$:} In Section~\ref{Hyperedge-Based Lead-Lag Relationships Extraction}, for time series at each scale, we study the lead-lag correlations within $w^s$ time steps to better extract complex dependencies in the time series, thereby improving forecasting performance. Here, we select three scales and specify the lead-lag time steps as $w^1$, $w^2$, and $w^3$, respectively. As shown in Figure~\ref{Lead-Lag steps}, we find that different datasets have variations in their multi-scale lead-lag time steps when achieving optimal performance. This is because different datasets possess unique characteristics, and their cyclicity (fine-grained scale) and trend (coarse-grained scale) may change due to external factors. Therefore, we recommend adjusting this parameter specifically when dealing with new datasets.

\textbf{Score weight $\alpha$:} Besides, we adopt the score weight in section~\ref{Optimization Objectives} to trade off the MSE loss and the ranking-aware loss---see Figure ~\ref{Score weight}. Experimental results show that all datasets achieve optimal performance and exhibit high stability when the score weight is set to 1. Therefore, we recommend setting this parameter to 1 in future research to obtain more optimized results.

% \begin{figure}[t]
%     \centering
%     \includegraphics[width=0.5\linewidth]{figures/lead-lag.pdf}
%     \caption{The specific lead-lag interaction process within the window, with a lead-lag step of 3 as an example.}
%     \label{The specific lead-lag interaction process within the lead-lag window.}
% \end{figure}

% \begin{figure}[t]
%     \centering
%     \includegraphics[width=0.6\linewidth]{figures/message.pdf}
%     \caption{Three types of message passing: hypernodes for stocks and hyperedges for industries.}
%     \label{Three types of message passing}
% \label{industry correlation.}
% \end{figure}

% \newpage
% \input{checklist.tex}

\end{document}